%% file: neurips_2022.tex
\documentclass{article}

    \PassOptionsToPackage{numbers, compress}{natbib}
\usepackage[preprint]{neurips_2022}

\usepackage[utf8]{inputenc} % allow utf-8 input
\usepackage[T1]{fontenc}    % use 8-bit T1 fonts
\usepackage{hyperref}       % hyperlinks
\usepackage{url}            % simple URL typesetting
\usepackage{booktabs}       % professional-quality tables
\usepackage{amsfonts}       % blackboard math symbols
\usepackage{nicefrac}       % compact symbols for 1/2, etc.
\usepackage{microtype}      % microtypography
\usepackage{xcolor}         % colors

\usepackage{subcaption}
\usepackage{graphicx}
\usepackage{caption}

\usepackage{amssymb}
\usepackage{amsmath}
\DeclareMathOperator*{\argmin}{arg\,min}
\DeclareMathOperator{\clip}{clip}
\DeclareMathOperator{\round}{round}
\DeclareMathOperator{\potwo}{PO2}
\DeclareMathOperator{\ceil}{ceil}
\DeclareMathOperator{\floor}{floor}
\usepackage{mathtools}

\usepackage{algorithm}
 \usepackage{algpseudocode}
 \usepackage{algorithmicx}
 \algdef{SE}[DOWHILE]{Do}{doWhile}{\algorithmicdo}[1]{\algorithmicwhile\ #1}%
\algrenewcommand\algorithmicrequire{\textbf{Input:}}
\algrenewcommand\algorithmicensure{\textbf{Output:}}

\usepackage{bbm}

\usepackage{sidecap}
\sidecaptionvpos{figure}{c}
\usepackage{authblk}

\title{A Closer Look at {\it{Hardware-Friendly}}\\Weight Quantization}
\author[]{Sungmin Bae}
\author[]{Piotr Zielinski}
\author[*]{Satrajit Chatterjee\thanks{This work was done when the author worked at Google, Mountain View, CA.}}
\affil[]{Google Research}
\affil[]{\texttt{\{smbae,zielinski\}@google.com}}
\affil[*]{\texttt{satrajit@gmail.com}}

\begin{document}
%\nolinenumbers
\maketitle
\input{body/body}

\bibliographystyle{plainnat}
\bibliography{references/paper.bib}
\newpage
% \input{body/checklist}
% \newpage
\appendix
\input{body/appendix}

\end{document}

%% file: body/body.tex
\begin{abstract}
\label{sec:abstract}

Quantizing a Deep Neural Network (DNN) model to be used on a custom accelerator with efficient fixed-point hardware implementations, requires satisfying many stringent hardware-friendly quantization constraints to train the model.
We evaluate the two main classes of hardware-friendly quantization methods in the context of weight quantization: the traditional Mean Squared Quantization Error (MSQE)-based methods and the more recent gradient-based methods. We study the two methods on MobileNetV1 and MobileNetV2 using multiple empirical metrics to identify the sources of performance differences between the two classes, namely, sensitivity to outliers and convergence instability of the quantizer scaling factor. Using those insights, we propose various techniques to improve the performance of both quantization methods - they fix the optimization instability issues present in the MSQE-based methods during quantization of MobileNet models and allow us to improve validation performance of the gradient-based methods by 4.0\% and 3.3\% for MobileNetV1 and MobileNetV2 on ImageNet respectively.
\end{abstract}

%% Introduction
\section{Introduction}
\label{sec:intro}
Recently there has been a rapid increase in demand for deploying Deep Neural Network  (DNN)-based inference tasks on battery-powered edge devices for privacy and latency requirements. A hardware-based DNN accelerator designed for an specific application is shown to be a promising approach to process such computationally challenging tasks under a tight power budget constraint, which is much faster and significantly power-efficient than a general purpose DNN accelerator~\cite{Wang2019}. In order to compress a DNN model small enough to fit it onto such an accelerator, quantization methods need to deal with more stringent constraints~\cite{habi2020hmq} such as uniform and symmetric quantization, power-of-2 (PO2) scaling factors, and batch normalization layer folding~\cite{ioffe2015batch}. The goal is to quantize model weights down to 4 bits or lower, but even with quantization-aware training, doing so without incurring a significant accuracy loss remains a challenging research problem for the community. 

There are two main classes of quantizer optimization methods for quantization aware training: the Mean Squared Quantization Error (\textbf{MSQE})\textbf{-based} methods~\cite{7178146, qkeras, rastegari2016xnornet, zhang2018lqnets, hou2018lossaware} and the more recent \textbf{gradient-based} methods~\cite{jain2020trained, jung2018learning, yang2019quantization, choi2018pact, esser2020learned, park2020profit}.
The \textbf{MSQE-based} methods iteratively search for {\it{optimal}} quantizer parameters at each training step minimizing the MSQE between the full-precision value and its quantized low-precision counterpart. There are also MSQE-based methods which formulate the optimization problem to directly minimize the loss as in~\cite{hou2018lossaware, hou2018lossaware_ternary}.
{\it{QKeras}}~\cite{qkeras}, a popular framework for hardware-friendly quantization aware-training, has extended the MSQE-based methods with the hardware-friendly constraints. 
The \textbf{gradient-based} methods optimize the quantization parameters directly during gradient descent steps, so they can be jointly learned with other network parameters to minimize the loss function. 
~\cite{choi2018pact} developed a method to learn a clipping parameter for activation function. ~\cite{esser2020learned} proposed a learnable scaling factor and means to estimate and scale their gradients to jointly train with other network parameters. 
~\cite{jain2020trained} suggested a gradient-based hardware-friendly quantizer\footnote{It is similar to LSQ quantizer in~\cite{esser2020learned} with the additional PO2 scaling factor constraint.} which optimizes power-of-2 (PO2) constrained clipping range\footnote{It is equivalent to learning the PO2 scaling factor for fixed bit-widths, uniform, and symmetric quantization.}.

In this paper we study MSQE- and gradient-based methods for weight quantization on commonly used vision backbones with the objective of understanding the sources (underlying causes) of performance difference between the two classes of method. Based on our insights, we propose improvements to the training process that increase stability of training and the performance of the final models to achieve validation accuracy of $66.9 \pm 0.17\%$ on MobileNetV1 and $67.5 \pm 0.01\%$ on MobileNetV2. To the best of our knowledge, we are not aware of any work in the literature that satisfies all our hardware-friendliness constraints. %~\footnote{For example, the recent work ~\cite{pmlr-v139-zhang21r} has a long list of quantization configurations from the literature in their Table 1, but to our knowledge none of these configurations simultaneously include folding or quantizing the batch norm layers and power-of-2 scaling, and many use more than 4 bits to quantize weights or activations.}.
The code for the quantizers will be available on Github~\footnote{https://github.com/google/qkeras/tree/master/experimental/quantizers}.

The rest of the paper is organized as follows. Section 2 provides background on hardware-friendly quantization and describes the two main classes of hardware-friendly quantizers. In Section 3, we analyze the effects of quantization during training of the quantizers, and we improve the performance of both quantizers based on the findings of the analysis. Finally, experiments and conclusion can be found in Section 4 and Section 5, respectively.
%
%% Background
\section{Background}
\label{sec:background}
In this section, we describe hardware-friendly quantization constraints for efficient custom hardware implementation and the two main classes of hardware-friendly quantization methods used in our study.  
\subsection{Quantization for Deep Neural Networks}
Quantization in DNN is a process of carefully searching for quantization parameters such as the scaling factor ($\Delta$) in~\eqref{eqn:quantization_simple} below\footnote{It models uniform and symmetric quantization, where $q_{\min}=-2^{bw-1}+1$, $q_{\max}=2^{bw-1}-1$ and  $q_{\min}=0$, $q_{\max}=2^{bw}-1$ for signed and unsigned input, respectively ($+1$ in the signed $q_{\min}$ is to match the number of quantization levels of $q_{\max}$, and $bw$ is the bit width of the quantizer.).} to map model parameters (e.g., weight, bias) and/or activations from a large set to approximated values in a finite and smaller set using a function $Q : \mathbb{R} \rightarrow C$ with minimum accuracy degradation as shown in~\eqref{eqn:quantization_simple}. 
\begin{equation}\label{eqn:quantization_simple}
    % \begin{align}\label{eqn:quantization_simple}
    w_q = Q(w, \Delta) = \Delta \cdot \clip(\round(\frac{w}{\Delta}); q_{\min}, q_{\max}),\; \text{where}\; w \in R, w_q \in C
    % \end{align}
\end{equation}
\subsection{Hardware-Friendly Quantizer Constraints}
\textbf{Uniform quantization}, which has uniformly spaced quantization levels, is often preferred due to its simplicity in hardware. Non-uniform quantization techniques like K-mean clustering based methods requires code-book look-up table related hardware overhead and logarithmic based methods can have a problem so called \emph{rigid resolution}\footnote{Having a larger bit-width only increases the quantization levels concentrated around 0.}~\cite{li2020additive}. Adding multiple logarithmic quantization terms was suggested to solve the problem~\cite{li2020additive}, however it incurs hardware overhead. 

\textbf{Symmetric quantization} has a symmetrical number of quantization levels around zero. It has less hardware overhead than asymmetric quantization as the zero point (i.e., quantization range offset) is 0. Even some terms related to the zero point (when it is non-zero) can be pre-computed and added to the bias, however there is a term unavoidably calculated on-the-fly with respect to the input~\cite{krishnamoorthi2018quantizing}.

\textbf{Per-tensor (layer) quantization} has a single scaling factor (and zero point) per entire tensor. It has better hardware efficiency than per-channel quantization, which has different scaling factor for each output channel of the tensor, since all the accumulators for entire tensor operate using the same scaling factor.

\textbf{Batch Normalization (BN) layer folding}~\cite{krishnamoorthi2018quantizing} reduces the two-step computations of the batch-normalized convolution to a one-step convolution by merging batch normalization operation with the weight and the bias of an adjacent linear layer~\cite{ioffe2015batch}.

\textbf{Power-of-2 (PO2) scaling factor} improves the hardware efficiency by constraining the scaling factor to a power-of-2 integer (that is to $2^s$ where $s$ is an integer) as then the scaling operation only requires a simple bit-shift operation instead of a fixed point multiplication. However, it often results in the accuracy degradation since the expressiveness of the scaling factor is limited.

\subsection{MSQE-based Hardware-Friendly Quantizer}
An MSQE-based hardware-friendly quantizer directly optimizes the PO2 scaling factor to minimize the mean squared quantization error~\cite{qkeras}, in its basic form, which is shown in~\eqref{eqn:msqe_opt}. 
\begin{align}
    \label{eqn:msqe_opt}
    {\Delta_{\potwo}^* = \argmin_{\Delta_{PO2}} || Q(w, \Delta_{PO2}) - w  ||^2}\;, \;\textrm{where} \;\Delta_{PO2} = PO2(\Delta) = 2^{\round(\log_2(\Delta))}
\end{align} 
Algorithm~\ref{alg:msqe-opt} describes a pseudo-code of the MSQE optimization. It reduce the quantization problem to a linear regression problem with a closed form solution to find an optimal scaling factor in terms of the minimum MSQE, where it iteratively performs least squares fits at each training step~\cite{zhang2018lqnets}. The initial scaling factor {$\Delta_{init}$} can be either the previously found scaling factor or a rough estimation based on the input distribution. We call it $\sf{MSQE}$ and use it as the baseline (and specifically the implementation from {\it{QKeras}}~\cite{coelho2021automatic} (i.e., {\it{quantized\_bits}})) for the MSQE-based hardware-friendly quantizer in our study. 
\begin{algorithm}
\caption{MSQE-based Optimization}\label{alg:msqe-opt}
    \begin{algorithmic}
        \Require $w, {\Delta_{init}}, {N}_{iters}$
        \Ensure ${\Delta_{PO2}}$
        \Procedure{:}{}
        \State $q \gets \frac{Q(w, \Delta_{init})}{\Delta_{init}}$
        \State $N \gets {N}_{iters}$
        \While{$N \neq 0$}
        \State $\Delta \gets \cfrac{q^{T}w}{q^{T}q}$ \Comment{\textit{Find an unconstrained optimal $\Delta$ minimizing the MSQE.}}
        \State $\Delta_{PO2} \gets PO2(\Delta)$ \Comment{\textit{Constrain $\Delta$ to be a power-of-2 value.}}
        \State $q \gets \frac{Q(w, \Delta_{PO2})}{\Delta_{PO2}}$ \Comment{\textit{Quantize w with the newly found  $\Delta_{PO2}$.}}
        \State $N \gets N - 1$
        \EndWhile
        \EndProcedure
    \end{algorithmic}
\end{algorithm}

\subsection{Gradient-based Hardware-Friendly Quantizer}
A gradient-based hardware-friendly quantizer learns the PO2 scaling factor from the gradients seeking to optimize the scaling factor to minimize the task loss directly instead of locally optimizing the MSQE objective at each training iteration~\cite{ esser2020learned, jain2020trained, Bhalgat2020LSQIL}.  
However, training $\Delta$ (in~\eqref{eqn:msqe_opt}) from the gradients can cause a numerical problem as the input to the $\log_2$ function can become negative by gradient update. It can be resolved by learning the scaling factor directly in the log domain ($\Delta_{\log_2}$) without $\log_2$ function~\cite{jain2020trained}\footnote{Softplus function can also be used to train $\Delta$ directly to avoid the numerical problem~\cite{park2020profit}.} as shown in~\eqref{eqn:neg-log2-domain}\footnote{~\cite{jain2020trained} suggests to use {\it{ceil}} function instead of {\it{round}} function to bias the scaling factor in the direction of having more elements within the quantization range, however {\it{round}} is denoted for simplicity.}.
% \footnote{We formulated a gradient-based hardware-friendly quantizer based on~\cite{jain2020trained} as the baseline, which is called $\sf{GRAD}$, in our study.}.
%
\begin{align}
\label{eqn:neg-log2-domain}
    \Delta_{PO2} = PO2(\Delta_{\log_2}) = 2^{{\rm round}(\Delta_{\log_2})}
\end{align}
 The local gradient of $\Delta_{\log_2}$ has dependency with the magnitude of the scaling factor ($2^{\Delta_{\log_2}}$) (see~\eqref{eqn:tqt_grad}\footnote{It uses a straight-through estimator (STE)~\cite{bengio2013estimating} in order to back-propagate through the round function.}), which can cause training instability (too large gradient) or slow convergence (too small gradient) depending on the scaling factor value~\cite{jain2020trained}. To lessen the problem, ~\cite{jain2020trained} suggests a method to normalize the gradient with its moving average variance and clip any large gradients for SGD training. However, it also reports that using Adam~\cite{kingma2017adam} optimizer without the normalization works well.
%\footnote{~\cite{jain2020trained} used Adam for their experiments.}.
%
\begin{align} 
\label{eqn:tqt_grad}
\begin{split}
    \frac{\partial w_q}{\partial \Delta_{\log_2}} = \frac{\partial w_q}{\partial \Delta_{PO2}} \cdot \frac{\partial \Delta_{PO2}}{\partial \Delta_{\log_2}}
    := \frac{\partial w_q}{\partial \Delta_{PO2}}\cdot 2^{\Delta_{\log_2}}\cdot \ln 2 
\end{split} 
\end{align}
The scaling factor can be initialized in several ways based on input values. It can be based on the dynamic range of the inputs, the input distributions~\cite{esser2020learned, jain2020trained}, and MSQE-optimal scaling factor~\cite{Bhalgat2020LSQIL}. We use a quantizer based on~\cite{jain2020trained} as the baseline for the gradient-based hardware-friendly quantizer, and call it $\sf{GRAD}$.

\section{Improving the Hardware-Friendly Quantizers}
We analyze the advantages and disadvantages of both quantizer types and improve their performance based on our analysis.
\subsection{Improving \texorpdfstring{$\sf{MSQE}$}{Lg}} \label{sec:improving_msqe}
\subsubsection{Sub-optimality in the Optimization of \texorpdfstring{$\sf{MSQE}$}{Lg}} \label{sec:line_search}
Algorithm~\ref{alg:msqe-opt} can converge to a sub-optimal solution as it makes a linear approximation to the original discrete non-linear problem. The sub-optimality can be improved by using a line search algorithm, where it simply searches for an MSQE-optimal PO2 scaling factor from the vicinity of the solution from Algorithm~\ref{alg:msqe-opt} (we will call the method as $\sf{finetune}$ in this paper.) (See Appendix~\ref{app:suboptimality} for an example and the method in detail.).

\subsubsection{A Problem with Outliers in \texorpdfstring{$\sf{MSQE}$}{Lg}} \label{sec:clipping}
\begin{figure}
\centering
  \includegraphics[width=.85\linewidth]{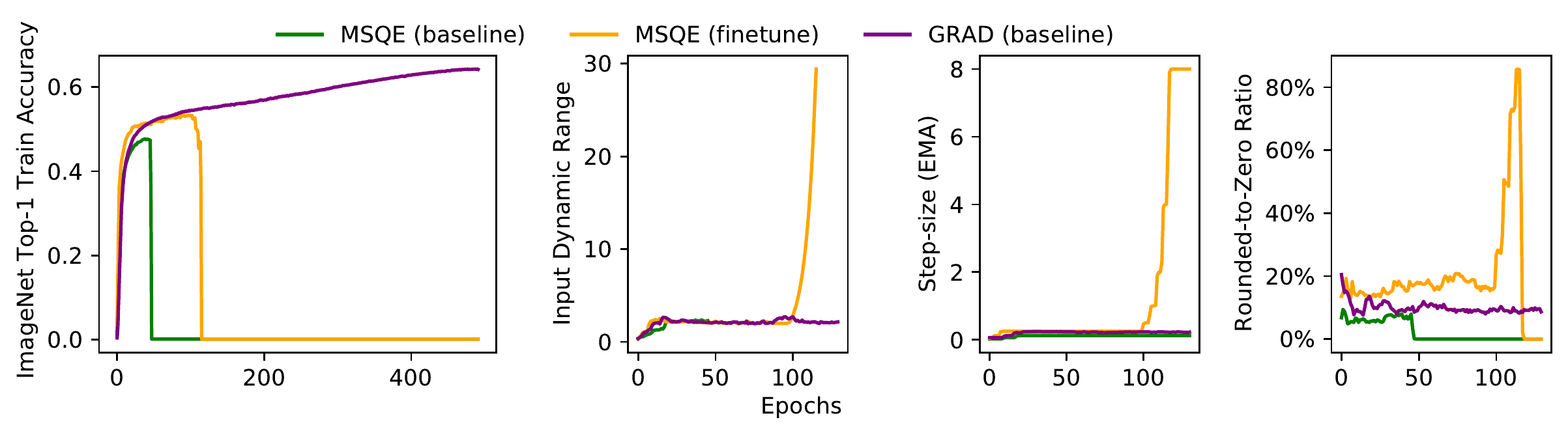}
\caption{$\sf{MSQE}$ and $\sf{MSQE} (finetune)$ were unstable and diverged due to outliers, while $\sf{GRAD}$ was stable, during training quantized MobileNetV1 (4W4A) from scratch on ImageNet.}%The various metrics from a layer (weight) show that the scaling factor of $\sf{MSQE} (finetune)$ followed rapidly increasing dynamic ranges of the input values which caused about 80$\%$ of the input values to round to zero. On the other hand, $\sf{GRAD}$ had well-controlled scaling factor in comparison. }
\label{fig:msqe_msqe_finetune_divergences}
\end{figure}

A quantizer should make {\it{good}} trade-off between rounding and clipping errors in order to keep the precision of the majority of the input values in the presence of outliers.
However, Algorithm~\ref{alg:msqe-opt} quantizes the input values uniformly between the largest and smallest values, so it is highly sensitive to outliers due to the squared error term in the optimization objective. This can be problematic especially for per-tensor quantization with batch normalization (BN) folding, where difference in the dynamic range of the weights between the channels can be large (see Figure~\ref{fig:app-max_min_dynamic_range_ratio} in the Appendix for details.).
%as shown in Figure~\ref{fig:max_min_dynamic_range_ratio}.
%
% \begin{figure}
% \centering
%   \includegraphics[width=.5\linewidth]{figs/quantizers/max_min_dynamic_range_ratio_mv1_msqe_finetune.pdf}
% \caption{It shows that the ratios between the maximum and the minimum dynamic range (within the channels of a layer) for the BN (Batch Normalization) folded weight and the BN unfolded weight, which were averaged across all the layers, from training MobileNetV1 with $\sf{MSQE} (finetune)$ quantizers. We can see that the BN folded weight generally had higher ratios compared to the BN unfolded one during training (about 15.9 times higher maximum dynamic range ratio).}
% \label{fig:max_min_dynamic_range_ratio}
% \end{figure}
%
Figure~\ref{fig:msqe_msqe_finetune_divergences} shows that the $\sf{MSQE}(baseline)$ and $\sf{MSQE} (finetune)$ were unstable and diverged, when $\sf{GRAD}$ was stable, during training quantized MobileNetV1 on ImageNet from scratch. The various metrics of a layer in Figure~\ref{fig:msqe_msqe_finetune_divergences} shows that the scaling factor of $\sf{MSQE} (finetune)$ followed the rapidly increasing dynamic ranges of the input (weight) values causing about 80$\%$ of them to round to zero, which leaded to training divergence. On the other hand, $\sf{GRAD}$ had a well-controlled scaling factor in comparison.
%
% \subsubsection{Outlier Suppressing in the MSQE Optimization} \label{sec:outlier_threshold}

\begin{SCfigure}
  \centering
  \includegraphics[scale=0.5]{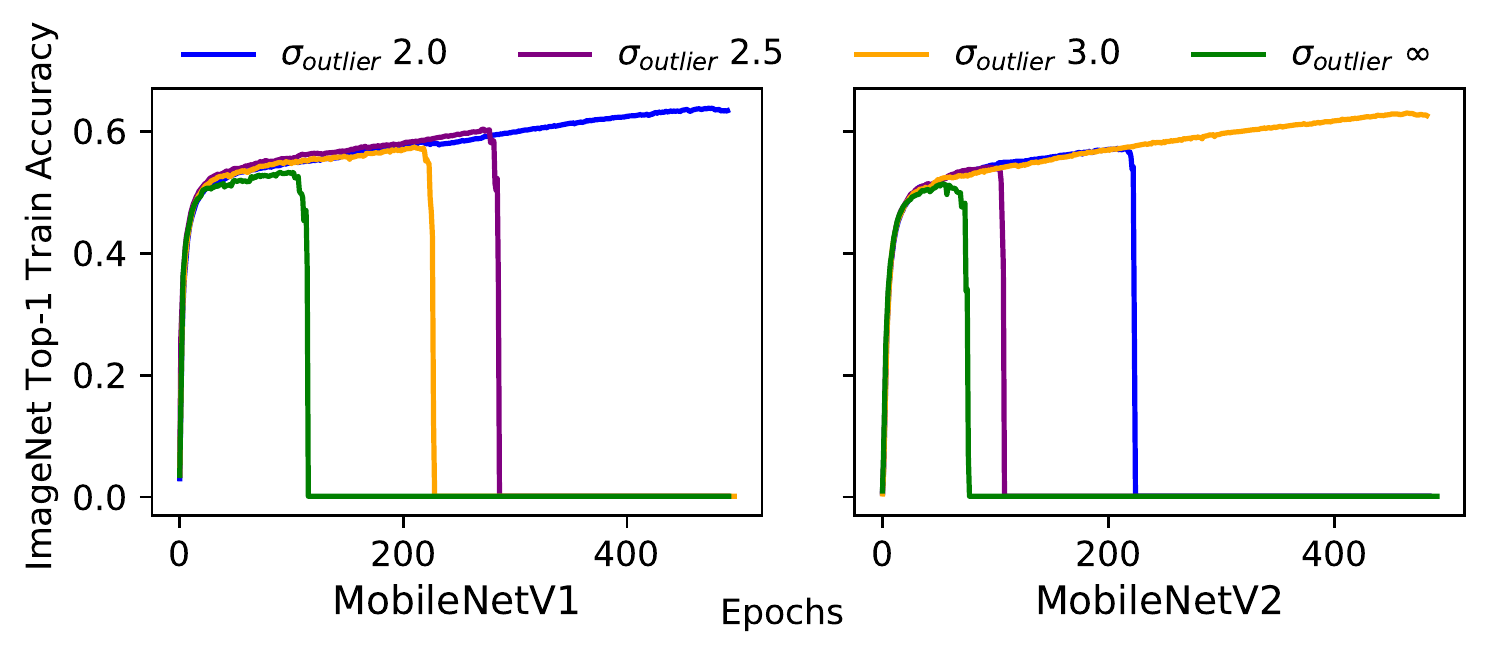}  
  \caption{The outlier method with the threshold of various standard deviations ($\sigma_{outlier}$ 2.0, 2.5, and 3.0) improves training stability compared to the one without the method ($\sigma_{outlier}$ $\infty$), but it does not solve the outlier problem entirely.}
  \label{fig:outlier_threshold}
\end{SCfigure}

To reduce the sensitivity to outliers, a weighting factor can be used to suppress them during the MSQE optimization at each training step by performing a weighted least squares fit for Algorithm~\ref{alg:msqe-opt} and a weighted line search (see the Appendix~\ref{app:outlier} for the method in detail.). ~\eqref{eqn:outlier_mask} shows a simple outlier mask ($M_{outlier}$) using a percentile based threshold ($\sigma_{outlier}$) (assuming the input ($w$) has a Gaussian distribution) to suppress outliers, where each element in the outlier mask gets either 1 or 0 depending on whether its corresponding (absolute) input value is less than the threshold or not.
\begin{equation}    
\begin{aligned}
\label{eqn:outlier_mask} 
% \State $M_{outlier} = \mathbbm{1}_{\sigma_{outlier}(w)}(|w|)$ \Comment{\textit{Element-wise 1 or 0, 1 if $w_{i} < T_{outlier}$, and 0 otherwise.}}
M_{outlier} = \begin{cases}  0& \text{if } |w| \geq \sigma_{outlier}\cdot \sigma(w) \\ 
                             1& \text{if } |w| <\sigma_{outlier}\cdot \sigma(w) \end{cases}
\end{aligned}
\end{equation}
Figure~\ref{fig:outlier_threshold} shows experiments using the outlier masking method with various standard deviations as the threshold. As we can see, the method didn't solve the problem entirely\footnote{The Gaussian distribution assumption may not be always valid.}, however it is clear that the training stability improved over the one without the method ($\sigma_{outlier}$ $\infty$).
% Replaced by the above figure with side caption
%  \begin{figure}[ht]
%     \centering
%     \includegraphics[scale=0.5]{figs/quantizers/outlier_thresholds_sig2.pdf}
%     \caption[$line_search$]{It shows experiments using the outlier masking method with 2.0 standard deviations as the threshold ($\sigma_{outlier}$ 2.0). It didn't solve the outlier problem entirely, however, it is clear that the training stability improved over the one without the method.}
%     \label{fig:outlier_threshold}
% \end{figure}
%
\subsubsection{Gradient Variance Awareness (GVA) Heuristic for Loss-Aware Optimization of \texorpdfstring{$\sf{MSQE}$}{Lg}}
Algorithm ~\ref{alg:msqe-opt} optimizes uniformly across all the input values without considering their $\it{importance}$ in the loss function. Similarly to~\cite{choi2017limit, hou2018lossaware_ternary, hou2018lossaware}, it can be improved by formulating the optimization objective as minimizing the second-order approximated loss degradation from quantization, which is often further simplified by ignoring the first-order term assuming the loss is near a local minimum, which is shown in~\eqref{eqn:second_order_loss}. 
\begin{equation}    
\begin{aligned}
\label{eqn:second_order_loss}    
    \argmin_{\Delta_{po2}} \mathbb{E}[L(w_{q}) - L(w)] &\approx \mathbb{E}[\frac{\partial L(w)}{\partial w}^T (w_{q} - w) + \frac{1}{2}(w_q - w)^T H(w) (w_q - w)] \\
    &\approx \mathbb{E}[\frac{1}{2}(w_q - w)^T H(w) (w_q - w)] \\
    where, &\; w_q=Q(w,\Delta_{po2}),\; H(w) = \frac{\partial^2 L(w)}{\partial w^2}
\end{aligned}
\end{equation}
In practice, the Hessian in~\eqref{eqn:second_order_loss} is approximated by a diagonal matrix~\cite{HAWQ2019}, which is sometimes further approximated by the (diagonal) empirical Fisher due to computational challenges for accurately estimating the diagonal entries of the Hessian~\cite{choi2017limit}. However, even the empirical Fisher approximation practically works well in some cases, conditions and assumptions that make the approximation valid are unlikely to be satisfied in practice~\cite{Kunstner2019}.
%
% In practice, the Hessian in the~\eqref{eqn:second_order_loss} is approximated by a diagonal matrix~\cite{HAWQ2019}, which is sometimes further approximated by the (diagonal) empirical Fisher due to computational challenges for accurately estimating the diagonal entries of the Hessian~\cite{choi2017limit}. However, even the empirical Fisher approximation practically works well in some cases, conditions and assumptions\footnote{The empirical Fisher converges to the true Fisher at a minimum when the model is good fit for the data with the assumptions that the model is able to realize the true data distribution and the number of data sample is sufficiently large~\cite{Kunstner2019}.} that make the approximation valid are unlikely to be satisfied in practice~\cite{Kunstner2019}.
%
For computational efficiency, the optimization problem solved in our study (see~\eqref{eqn:second_moment_quantization}), we use the (diagonal) empirical Fisher approximation ($v_j$ is a running average of the second (raw) order moments of the j-th weight element ($w_j$) of a layer) as a gradient variance metric to minimize the quantization error for more dynamically changing weights during training, which is similar to the gradient noise adaptation perspective in~\cite{Kunstner2019} (we call the method as gradient variance aware (GVA) optimization in our study.). 
\begin{equation}    
\begin{aligned}
\label{eqn:second_moment_quantization}  
    \argmin_{\Delta_{PO2}}&\;\frac{1}{2}\sum_{j=1}^{N_w} v_j\cdot (Q(w_j, \Delta_{PO2}) - w_j)^2 
\end{aligned}    
\end{equation}

\begin{figure}[ht]
    \centering
    \includegraphics[width=.35\linewidth]{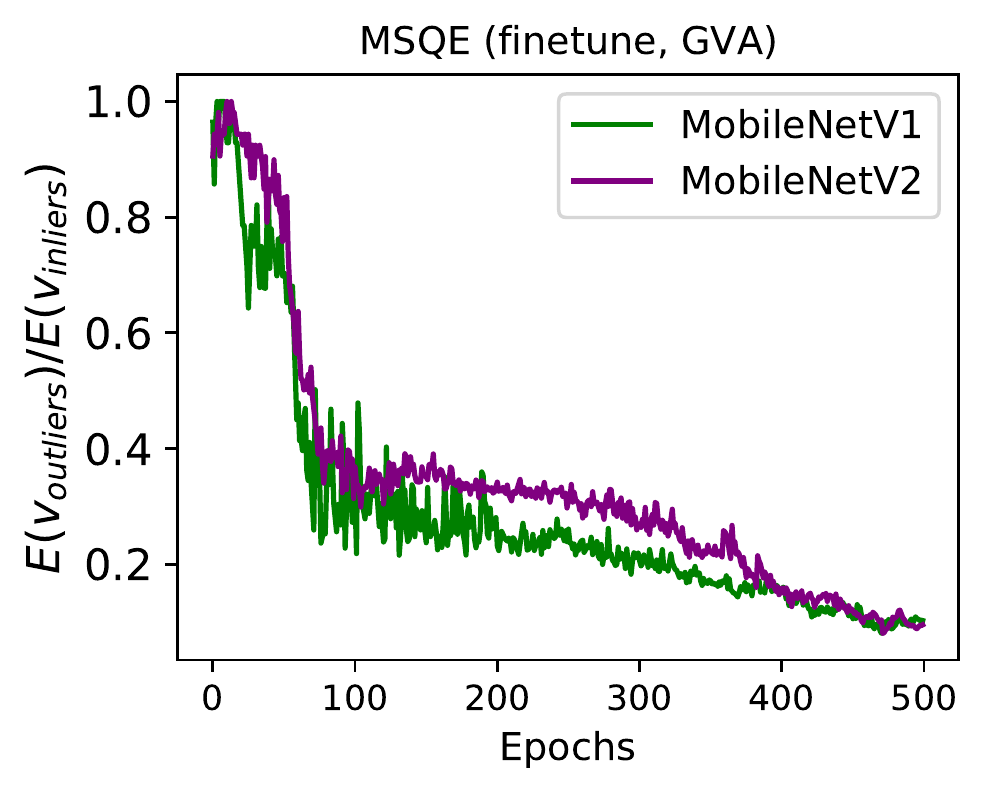} 
    \includegraphics[width=.52\linewidth]{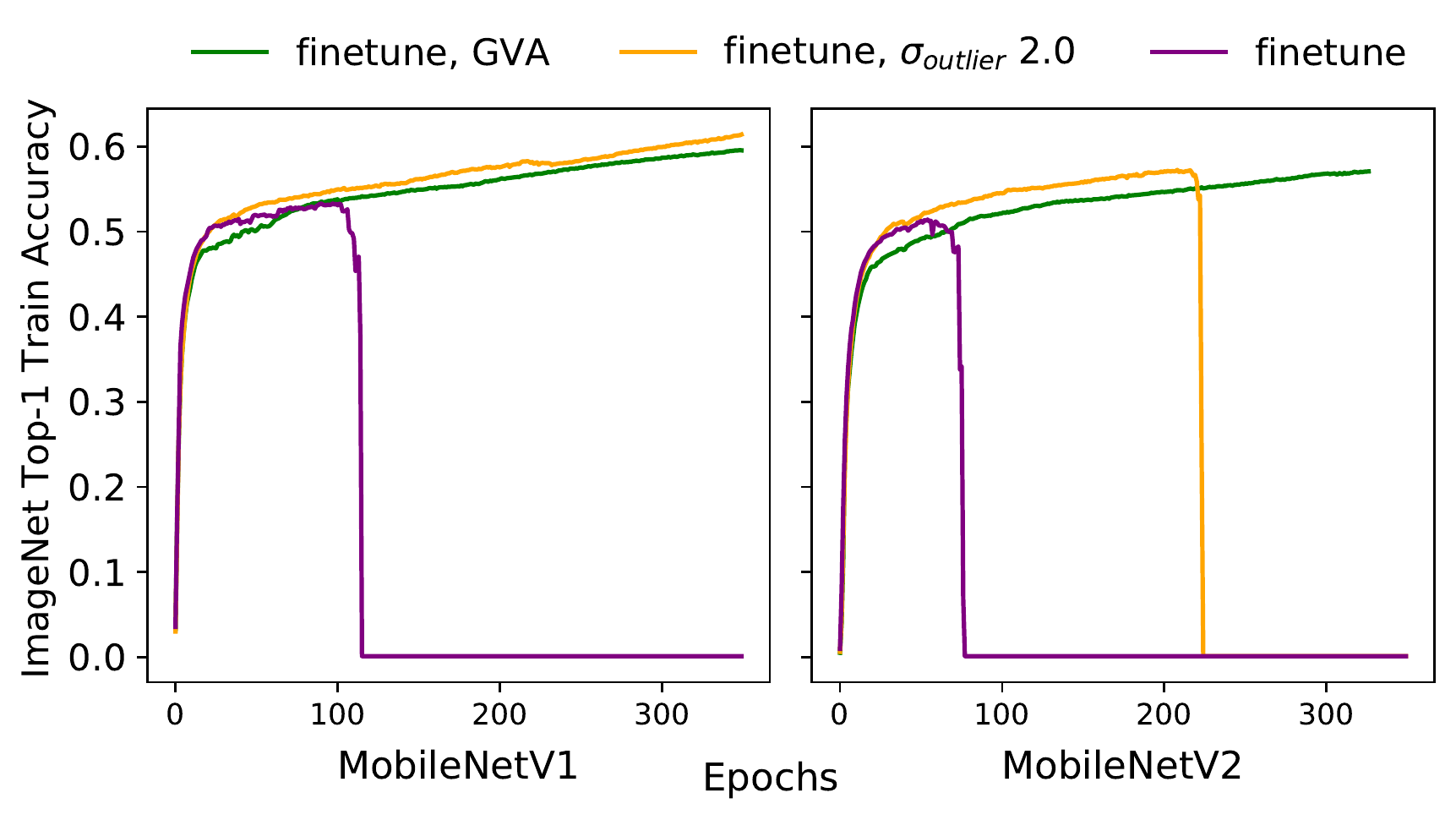}
    \caption[$line_search$] {The gradient variance aware (GVA) method provides an additional benefit of de-emphasizing outliers during training, where the average second order moments ratios between the outliers and the inliers become smaller and are less than 1. MobileNetV1 and MobileNetV2 (4W4A) were able to be trained from scratch without the outlier method.} \label{fig:second_moment_ratio}
\end{figure}

The method provides an additional benefit of suppressing outliers during training, which can complement the outlier method, since the running-averaged second (raw) order moments of the clipped inputs gradually decrease as long as they are clipped\footnote{The local gradients of the clipped inputs are zero.}. Figure~\ref{fig:second_moment_ratio} (left) shows that the average second (raw) order moment ratio between the outliers and the inliers of weight (averaged across all the layers) for training quantized MobileNetV1 and MobileNetV2 with $\sf{MSQE}$ with $\sf{finetune}$ and GVA ($\sf{MSQE}(finetune, GVA)$). The ratios become smaller as the models are trained longer which means that outlier weights tend to be de-emphasized for the MSQE optimization when the GVA method is used. Figure~\ref{fig:second_moment_ratio} (right) also shows the effectiveness of the GVA method that the models were able to be trained from scratch without the outlier mask method. 

\begin{figure}[ht]
    \centering
    \includegraphics[width=.8\linewidth]{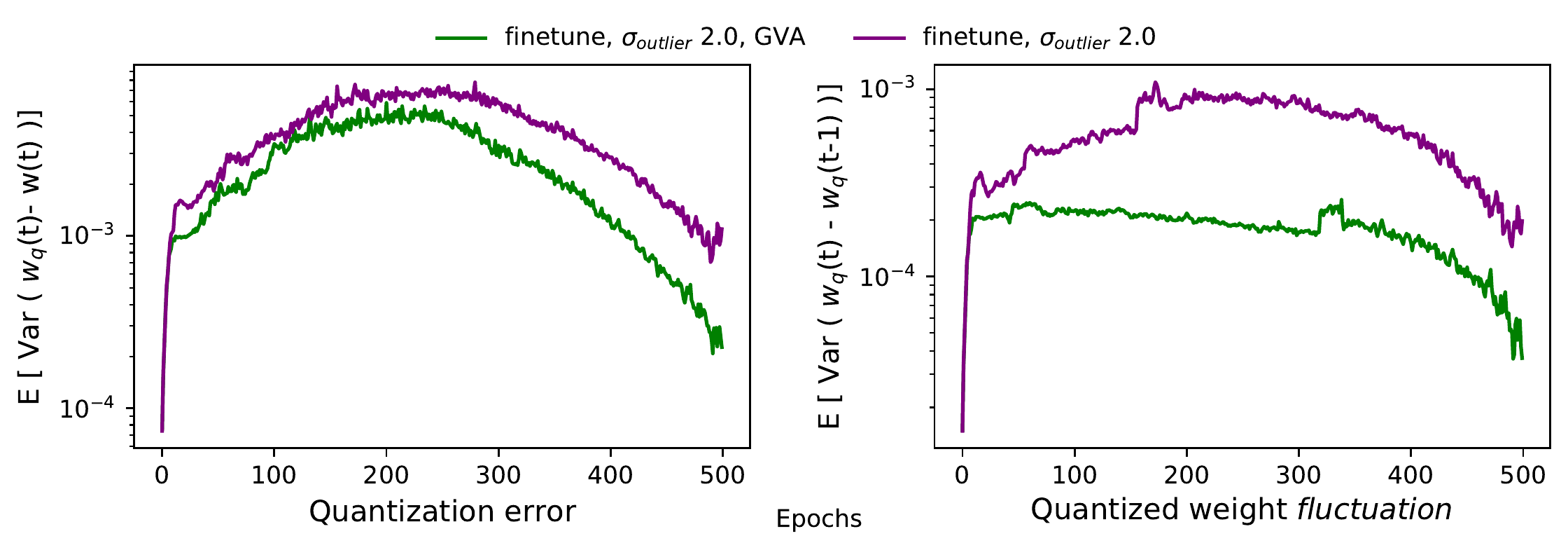}
    \caption[$line_search$]{Gradient variance aware optimization (GVA) reduces the variance in average quantization error (left figure) and the variance in average quantized weight $\it{fluctuation}$ (right figure) on MobileNetV1. The metrics were measured from the weights and averaged across all the layers.}
    \label{fig:qnoise_and_fluctuation_mv1}
\end{figure}

Figure~\ref{fig:qnoise_and_fluctuation_mv1} shows reduction in the average quantization error variance (left figure) and the average quantized weight $\it{fluctuation}$\footnote{Quantized weight $\it{fluctuation}$ means the difference between the current quantized weight and the one from the previous training step (i.e., $w_q(t) - w_q(t-1)$).} variance (right figure) by using the GVA method for MobileNetV1\footnote{The metrics were measured from the weights and averaged across all the layers.} (see the Appendix~\ref{sec:app-GVA_opt} for MobileNetV2 results.).

\subsection{Improving \texorpdfstring{$\sf{GRAD}$}{Lg}}
\subsubsection{Convergence Instability of \texorpdfstring{$\sf{GRAD}$}{Lg}}
\begin{SCfigure}
  \centering
  \includegraphics[scale=0.4]{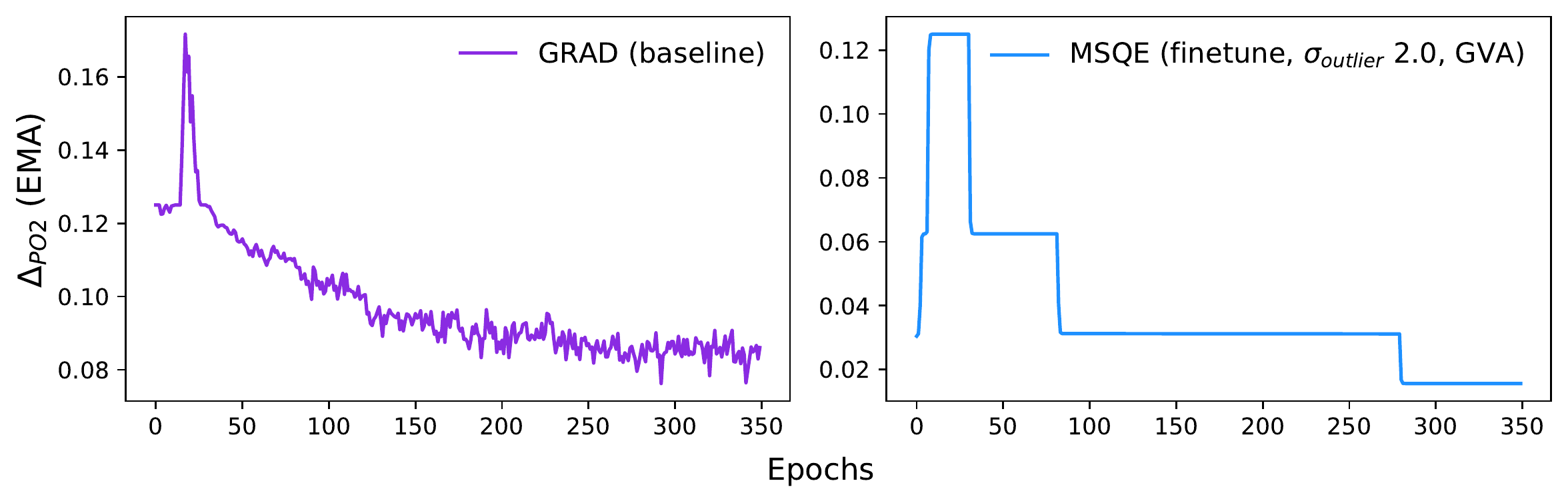}  
  \caption{The running averages of PO2 scaling factors of $\sf{GRAD}$ and $\sf{MSQE}$ $(\sf{finetune},$ $\sigma_{outlier}$ $2.0,$ $\sf{GVA})$ of a MobileNetV1 layer.}
  \label{fig:msqe_vs_grad_step_size_ema}
\end{SCfigure}

\begin{figure}[ht]
    \centering
    \includegraphics[scale=0.22]{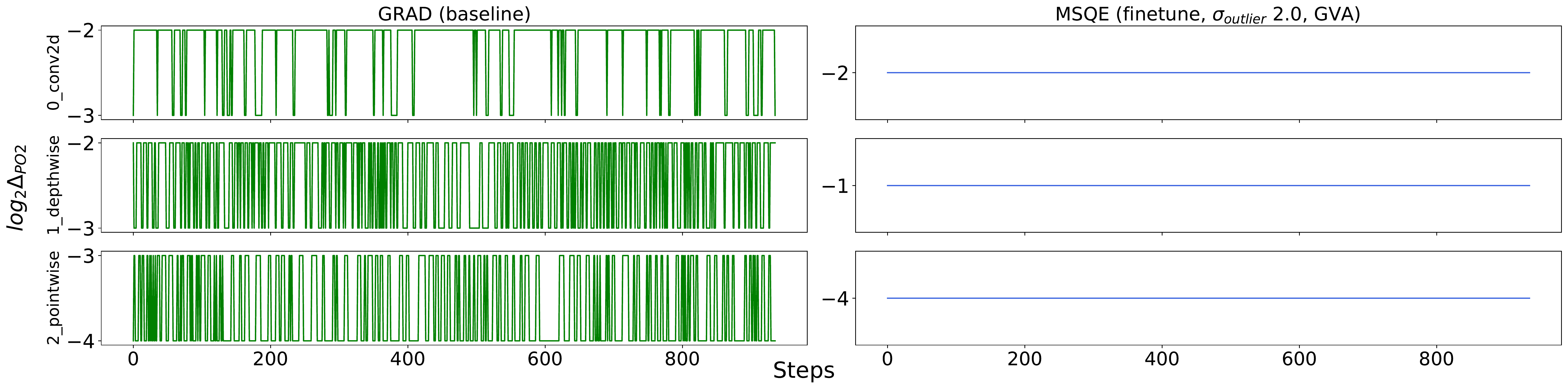}
    \caption[$msqe_vs_grad_train_stability$]{The (per-training step) PO2 scaling factor exponents of $\sf{GRAD}$ and $\sf{MSQE}$ $(\sf{finetune},$ $\sigma_{outlier} 2.0,$ $\sf{GVA})$ in the first 3 layers of MobileNetV1 for 3 epochs starting from the 150th epoch.}
    \label{fig:msqe_vs_grad_scale_fluctuation_mv1_150}
\end{figure}

Stably training the gradient-based hardware-friendly quantizer requires considerations for scaling factor convergence~\cite{jain2020trained}. Figure~\ref{fig:msqe_vs_grad_step_size_ema} shows running-averaged PO2 scaling factors of $\sf{GRAD}$ and $\sf{MSQE}$ $(\sf{finetune},\sigma_{outlier} 2.0, GVA)$ from a layer of MobileNetV1\footnote{They are running averages (exponential decay of 0.99) of scaling factors from a layer (MobileNetV1) trained with Adam~\cite{kingma2017adam} optimizer and cosine learning rate decay.}. We can see that there was more fluctuation in the PO2 scaling factor ($\Delta_{PO2}$) for $\sf{GRAD}$ than $\sf{MSQE}$ $(\sf{finetune},\sigma_{outlier} 2.0, GVA)$. It is more evidently shown in Figure~\ref{fig:msqe_vs_grad_scale_fluctuation_mv1_150}, which captured (per training step) the PO2 scaling factor exponent ($\log_2 \Delta_{PO2})$ of both quantizers.
% \begin{figure}[ht]
%     \centering
%     \includegraphics[scale=0.4]{figs/quantizers/msqe_vs_grad_step_size_ema.pdf}
%     \caption[$msqe_vs_grad_step_size_ema$]{It shows running averages (with exponential decay 0.99) of the PO2 scaling factors of $\sf{GRAD}$ and $\sf{MSQE}$ of a MobileNetV1 layer. }
%     \label{fig:msqe_vs_grad_step_size_ema}
% \end{figure}
%
\paragraph{Scaling factor oscillation around the ceil (round) transition boundary}
As shown in Figure~\ref{fig:msqe_vs_grad_scale_fluctuation_mv1_150}, the PO2 scaling factor fluctuates between the two adjacent PO2 scaling factors when the scaling factor exponent ($\Delta_{\log_2}$) oscillates around the transition boundary of the ceil (round) function (see~\eqref{eqn:neg-log2-domain}). To improve this type of instability, we adopt the idea from $\sf{MSQE}$ that uses an MSQE-optimal PO2 scaling factor instead of statically performing ceil (round) function. For example, if we choose to round to a lower MSQE PO2 scaling factor between the adjacent PO2 scaling factors, it can reduce the fluctuation when the (unconstrained) scaling factor ($2^{\Delta_{\log_2}}$) is near a locally MSQE minimum PO2 scaling factor as shown in Figure~\ref{fig:round_to_lower_msqe} (we call the method $\sf{RTLM}$ ({\it{Round-to-Lower-MSQE}}) (see Appendix~\ref{app:conv_instability} for detailed algorithm). 
 \begin{figure}[ht]
    \centering
    \includegraphics[scale=0.27]{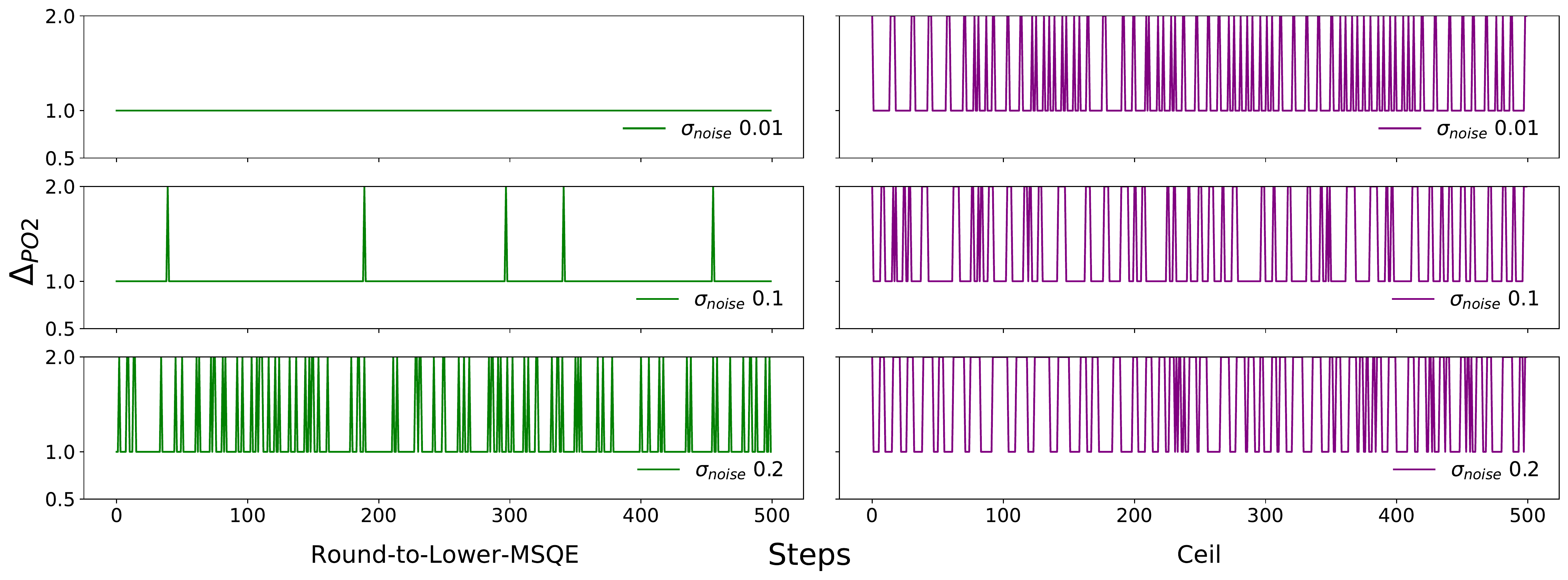}
    \caption[$round-to-lower-msqe$]{The effectiveness of $\sf{RTLM}$ ({\it{Round-to-Lower-MSQE}}) in improving the convergence instability to various random Gaussian noise perturbations to the input ($w$) for a toy example. It is a single quantizer configured to minimize the L2 error loss of $(w - Q(w,\Delta_{PO2}))^2$ using Adam~\cite{kingma2017adam} optimizer, where the (unconstrained) scaling factor ($2^{\Delta_{\log_2}}$) is initialized to near a locally MSQE minimum PO2 scaling factor ($1.0$) and the transition boundary of the ceil function.}
    \label{fig:round_to_lower_msqe}
\end{figure}
\paragraph{Scaling factor oscillation at convergence}
There is another type of convergence instability for $\sf{GRAD}$. Since the scaling factor is constrained to be a power-of-2 integer value, it cannot remain converged when an optimal scaling factor for a given input is not an exact power-of-2 integer value~\cite{jain2020trained} as shown in Figure~\ref{fig:step_size_fluctuation_at_convergence}. This can be a problematic especially near the end of training, where the weights are nearly fixed. It can be solved by freezing the scaling factor and using a running-averaged PO2 scaling factor instead at convergence\footnote{We believe it can be further improved by freezing the weights in a sophisticated sequence as in~\cite{park2020profit} after some epochs of fine-tuning with the frozen scaling factors.} (see Appendix~\ref{app:oscillation_at_convergence} for the method in details.).
 \begin{figure}[ht]
    \centering
    \includegraphics[scale=0.25]{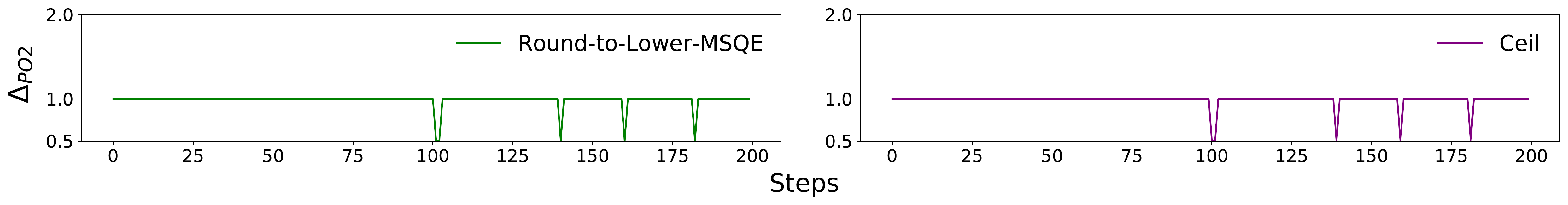}
    \caption[$scaling factor-fluctuation$]{The PO2 scaling factors for both methods periodically fluctuate between 0.5 and 1 when the optimal (unconstrained) scaling factor for the loss function is about 0.9 for a given fixed input (it used the same settings as in Figure~\ref{fig:round_to_lower_msqe}).}
    \label{fig:step_size_fluctuation_at_convergence}
\end{figure}
\section{Experimental Results}
We evaluate the MSQE-based and the gradient-based quantizers on MobileNetV1~\cite{howard2017mobilenets} and MobileNetV2~\cite{sandler2019mobilenetv2} which are one of the most popular backbone architectures for custom ML hardwares~\footnote{The code will be available https://github.com/google/qkeras/experimental/quantizers\_po2.py}.
\subsection{Experimental Settings}
We compare our $\sf{MSQE}$ improvements with the baseline implementation from QKeras \cite{qkeras}. \cite{jain2020trained} is our baseline for $\sf{GRAD}$.
MobileNetV1 and MobileNetV2 are quantized into 4 bits weight and 4 bits activation (including the first and the last layers) and 8 bits bias. (The bias is quantized into more bits than weights because it is a small fraction of the parameters in the network, and keeping higher precision for bias is important since each bias affects many output activations~\cite{jacob2017quantization, Lyon1996}.) In all experiments we use $\sf{GRAD}$ to quantize the activation outputs.
We ran each experiment 3 times and show the median run on the plots (wrt validation accuracy) as well as standard deviation ($\pm$) of results.
% \begin{table}
% \begin{center}
% \begin{tabular}{||c c c c c c c||} 
%  \hline
%  Weight & Bias & Activation & scaling factor & Granularity & BatchNorm & First (last) layers \\ 
%  \hline\hline
%  4 bits & 8 bits & 4 bits & PO2 & Per-layer & Folded & Quantized the same\\
%   & & & &quantization & & as other layers\\
%  \hline
% \end{tabular}
% \caption{\label{tab:quant_cofigs}Hardware-friendly quantization configurations}
% \end{center}
% \end{table}

The models were trained from scratch on the ImageNet~\cite{russakovsky2015imagenet} dataset to see their quantization characteristics during entire training process. It was optimized using Adam~\cite{kingma2017adam} optimizer, the batch of 4096\footnote{We used synchronized batch normalization to get the global moving batch statistics, otherwise, each TPU worker will train its own folded weights.}, the initial learning rate of 0.01, and cosine learning rate decay~\cite{loshchilov2017sgdr} with alpha of 0.001 and no warm restart, which were trained on TPU for 500 epochs. The training was regularized by random crop and random brightness saturation.
%
% \subsection{Evaluation of the Gradient Variance Awareness (GVA) Heuristic for $\sf{MSQE}$} \label{sec:gva_results}
\subsection{Evaluation of the Gradient Variance Awareness (GVA) Heuristic for $\sf{MSQE}$} \label{sec:gva_results}
\begin{figure}[ht]
    \centering
    \includegraphics[scale=0.42]{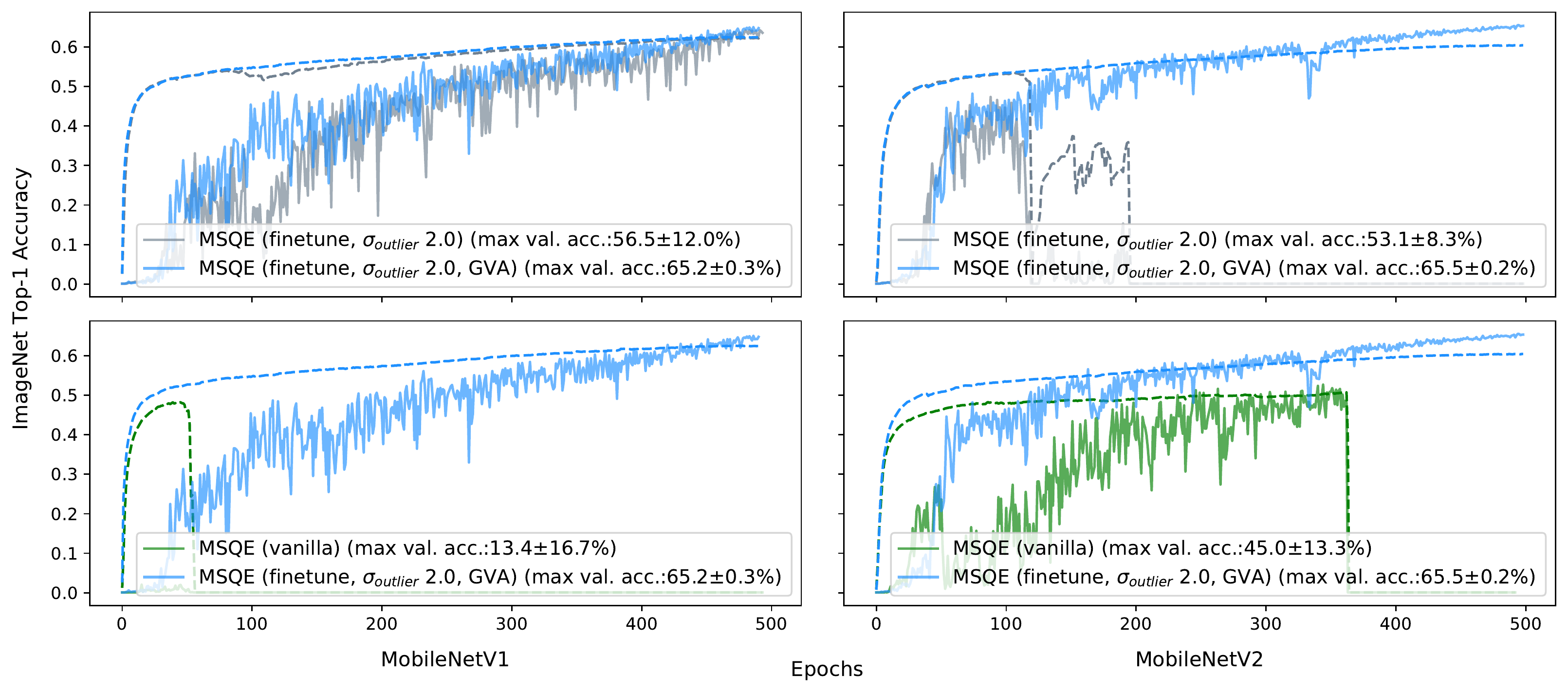}
    \caption[$scaling factor-fluctuation$]{(Top figures) Adding gradient variance awareness (GVA) results in better training accuracy (solid line), better validation accuracy (dotted line), and greater robustness (lower standard deviation at end of training). (Bottom figures) We can see that the optimization instability issues in $\sf{MSQE}$ were fixed by the techniques that we presented. We show the median run (out of 3).}
    \label{fig:outlier_gva}
\end{figure}

We performed an experiment to see how much the presence of the GVA method can affect the overall performance. As we can see from Figure~\ref{fig:outlier_gva}, adding the GVA method to the outlier mask only methods improved overall training stability and the validation accuracies (see the Appendix~\ref{sec:app-results-GVA} for training curves in detail and an additional ablation study.).
Figure~\ref{fig:performance_improvement} shows the overall performance improvements over the baseline MSQE quantization (even without the fine-tuning line search). Applying all the proposed techniques for the MSQE quantization  \textbf{made the optimization problem stable} without any other adjustments or regularizers, allowing it to reach over 65\% accuracy in both models.

% \begin{figure}
% \centering
%      {\includegraphics[scale=0.5]{figs/experiments/experimental_results_msqe_vanilla.pdf}} \\
%      \caption{The performance improvements from the vanilla \sf{MSQE}.}
%      \label{fig:msqe_vanilla}
% \end{figure}

\subsection{Evaluation of RTLM for $\sf{GRAD}$}
\begin{figure}[ht]
    \centering
    \includegraphics[scale=0.433]{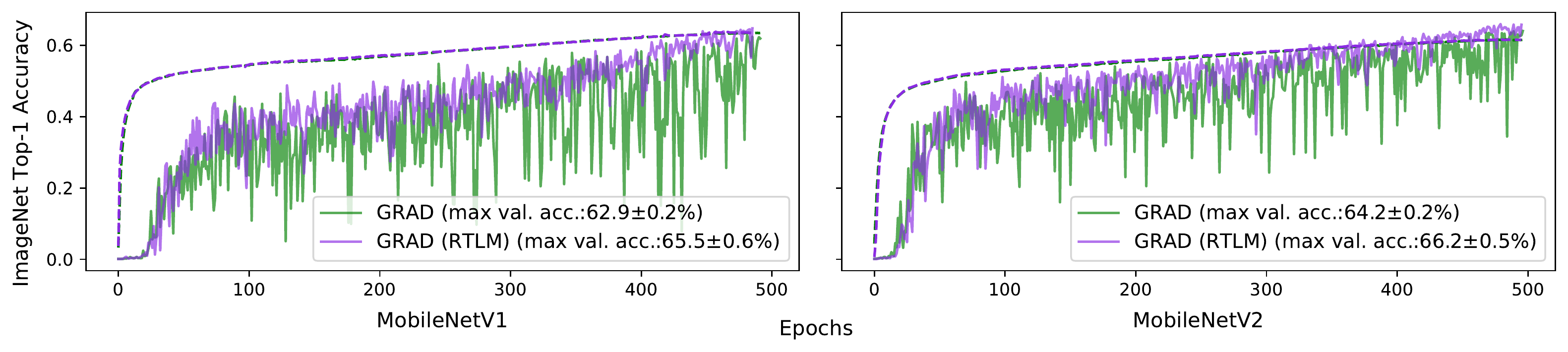}
    \caption[$scaling factor-fluctuation$]{It shows the effectiveness of RTLM method in improving training stability (based on the validation accuracy (solid line) fluctuations). Validation accuracy improved by $2.6\%$ and $2.0\%$ for MobileNetV1 and MobileNetV2, respectively.}
    \label{fig:performance_improvement}
\end{figure}

Figure~\ref{fig:performance_improvement} shows the effectiveness of RTLM method in improving training stability (based on the validation accuracy fluctuations). This can be explained by Figure~\ref{fig:average_scale_fluctuation} (in Appendix~\ref{sec:app-results-RTLM}) that measured the amount of scaling factor exponent changes (per training step) averaged across all the layers for 3 epochs from the corresponding epochs on the x-axis. We can see that RTLM had less average scaling factor fluctuations compared to the ceil function. %Figure~\ref{fig:scale_fluctuation_5layers} shows the scaling factor (exponent) changes of the first 5 layers of MobileNetV1 at the 450th epoch.
%
% \begin{figure}[ht]
%     \centering
%     \includegraphics[scale=0.37]{figs/experiments/average_scale_fluctuation_short.pdf}
%     \caption[$round-to-lower-msqe$]{It shows average scaling factor exponent fluctuations (per training step) for training MobileNetV1 and MobileNetV2 for 3 epochs from the corresponding epochs on the x-axis. We can see that RTLM had less average scaling factor (exponent) fluctuations compared to the ceil function.}
%     \label{fig:average_scale_fluctuation}
% \end{figure}
%
%  \begin{figure}[ht]
%     \centering
%     \includegraphics[scale=0.1]{figs/experiments/mv1_scale_fluctuation_5_layer_450.pdf}
%     \caption[$round-to-lower-msqe$]{It shows scaling factor exponent fluctuations (per training step) of Round-to-Lower-MSQE and Ceil functions for the first 5 layers of MobileNetV1 for 3 epochs from the 450th epoch.}
%     \label{fig:scale_fluctuation_5layers}
% \end{figure}
%
\subsection{Evaluation of Freezing the Scaling Factor at Convergence for $\sf{GRAD}$}
Freezing the scaling factor and using the running averaged scaling factor at convergence (at the 470th epoch) and fine-tuning the weights (for 30 epochs) further improved the validation accuracies for MobileNetV1 (MobileNetV2) for $\sf{GRAD}$ and $\sf{MSQE}$\footnote{The $\sf{MSQE}$-based models used $\sf{GRAD}$ quantizers for quantizing the activation output.} to  $66.9\pm0.17\%$ ($67.5\pm0.01\%$) and $65.6\pm0.18\%$ ($66.5\pm0.24\%$) respectively.
% output.} about 1.41$\%$ (1.30$\%$) and 0.42$\%$ (0.97$\%$) respectively.
%(see Appendix~\ref{app:oscillation_at_convergence} for details.).
%
% \begin{figure}
% \centering
%      \subfloat[][$\sf{GRAD}$ ]{\includegraphics[scale=0.5]{figs/experiments/experimental_results_freeze_step_size_grad.pdf}\label{fig:freeze_step_size_grad}} \\
%      %\subfloat[][$\sf{MSQE}$ ]{\includegraphics[scale=0.5]{figs/experiments/experimental_results_freeze_step_size_msqe.pdf}\label{fig:freeze_step_size_msqe}}
%      \caption{TODO: replace with table. Freezing the scaling factor (including the activations) for \sf{GRAD}\label{fig:freeze_step_size_grad} and \sf{MSQE}\label{fig:freeze_step_size_msqe} from the 470th epoch improved the validation accuracies for MobileNetV1 and MobileNetV2.}
%      \label{fig:freeze_step_size}
% \end{figure}
%
\section{Conclusions}
In this paper, we evaluated the two main classes of hardware-friendly quantization methods in the context of weight quantization: the traditional MSQE (Mean Squared Quantization Error)-based methods~\cite{qkeras} and the more recent gradient-based methods~\cite{jain2020trained}. We studied the two methods using multiple empirical metrics to  identify the sources (underlying causes) of performance difference between the two classes, and found that the outliers in the input and the scaling factor convergence instabilities are main problems for the MSQE-based and the gradient-based methods respectively. Using our insights, we proposed various techniques to improve the performance of both quantizers, which provided stability of the MSQE-based training and increased the validation accuracy of the gradient-based methods by 4.0\% and 3.3\% for MobileNetV1~\cite{howard2017mobilenets} and MobileNetV2~\cite{sandler2019mobilenetv2} on ImageNet~\cite{russakovsky2015imagenet} respectively. We believe that our insights could also be useful for activation quantization but leave that for future work.

%% file: body/appendix.tex
\part*{Appendices}\label{appendix}
%
% \section{Details on Hardware-Friendly Quantization Constraints}
% \textbf{Batch normalization folding}~\cite{krishnamoorthi2018quantizing} is a technique to fuse batch normalization~\cite{ioffe2015batch} operation with the weights and biases of a adjacent linear layer, which reduces the two-step computations of the batch-normalized convolution to a simple one-step convolution (see the equation~\eqref{eqn:app-bn_folding}).
% %
% \begin{equation}
% \begin{aligned}
% % \begin{gather}
%     \label{eqn:app-bn_folding}
%     y &= w^{\prime}\cdot x + b^{\prime} \\
%     w^{\prime} &= w\cdot \frac{\gamma}{\sqrt{\sigma_{batch}^{2}+\epsilon}} \\
%     b^{\prime} &= \frac{\gamma}{\sqrt{\sigma_{batch}^{2}+\epsilon}} \cdot (b - \mu_{batch}) + \beta 
% % \end{gather}  
% \end{aligned}
% \end{equation}
%
% To reduce jitter in the quantized weight due to noisy batch statistics ($\sigma_{batch}$), exponential moving average (EMA) statistics ($\sigma_{EMA}$) can be instead used to stabilize the training by applying batch normalization correction ($\frac{\sigma_{EMA}}{\sigma_{batch}}$) as shown in the \eqref{eqn:bn_folding_train}~\cite{krishnamoorthi2018quantizing}.
% %
% \begin{equation}
% \begin{aligned}
% % \begin{gather}
%     \label{eqn:bn_folding_train}
%     y &= \frac{\sigma_{EMA}}{\sigma_{batch}}\cdot (w_{q_{EMA}}^{\prime}\cdot x)  + b_{q_{batch}}^{\prime} \\
%     w_{q_{EMA}}^{\prime} &= Q(w\cdot \frac{\gamma}{\sqrt{\sigma_{EMA}^{2}+\epsilon}}) \\
%     b_{q_{batch}}^{\prime} &= Q(b^{\prime}) \\
% % \end{gather} 
% \end{aligned}
% \end{equation}
%
\section{Details on Sub-optimality in the Optimization of $\sf{MSQE}$} \label{app:suboptimality}
\begin{figure}[h]
    \centering
    \includegraphics[scale=0.5, trim={0 0.55cm 0 0}, clip]{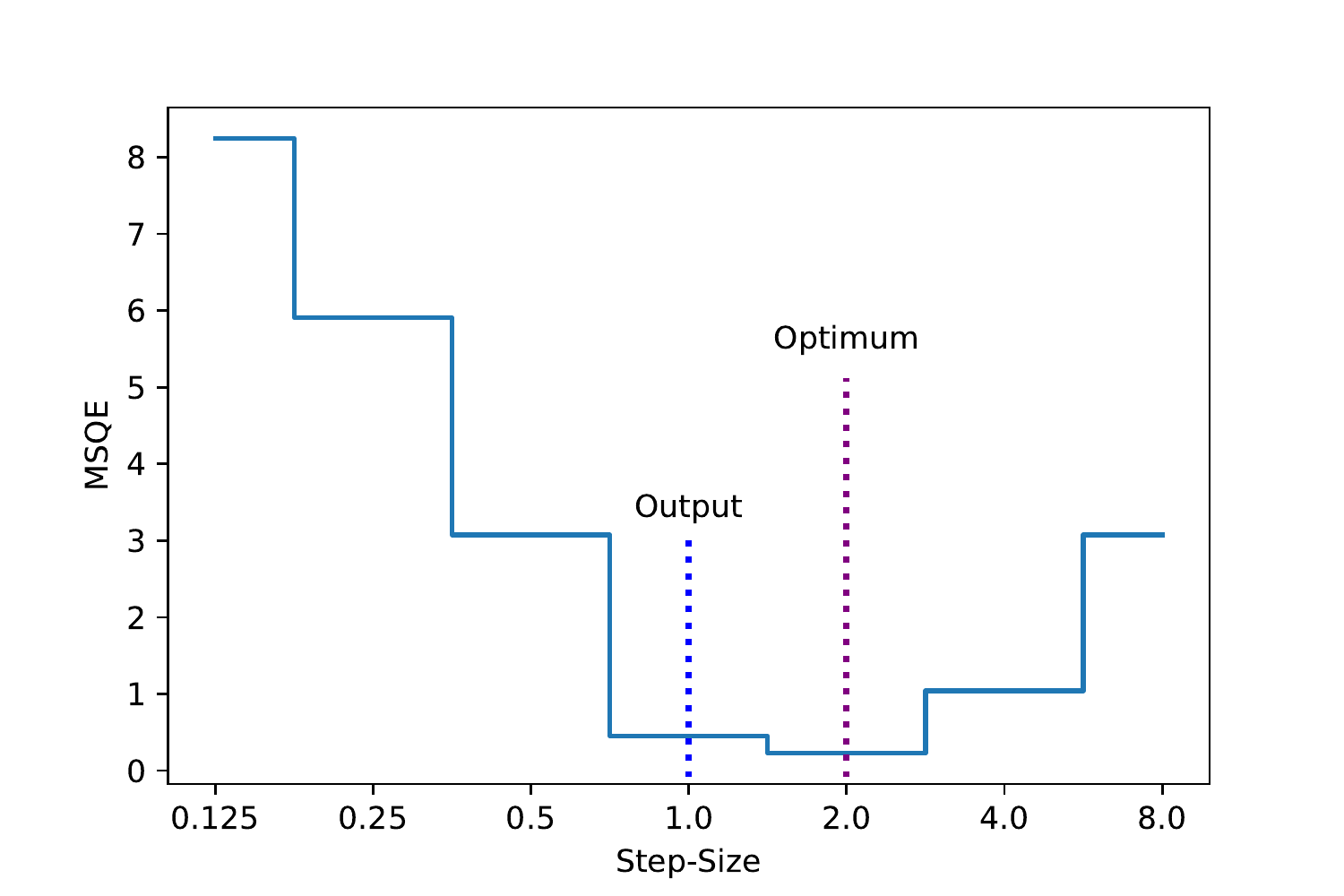}
    \caption[line_search]{It shows the MSQE for various PO2 scaling factors (x-axis) from the vicinity of the output (1.0) of Algorithm~\ref{alg:msqe-opt} for an example in~\eqref{eqn:MSQE_example}. We can see that 1.0 is not an MSQE-optimal PO2 scaling factor.}
    \label{fig:line_search}
\end{figure}

As shown in the example below (signed 4 bit quantization), it converges to a sub-optimal PO2 scaling factor of 1.0 for the initial scaling factor ($\Delta_{init}$) of 1.0, where 2.0 is an MSQE-optimal PO2 scaling factor (see figure~\ref{fig:line_search}). This sub-optimality can be improved by using a line search algorithm as in Algorithm~\ref{alg:line-search}, where it simply searches for an MSQE-optimal PO2 scaling factor from the vicinity of the solution from Algorithm~\ref{alg:msqe-opt}.
\begin{equation}
\begin{aligned} 
\label{eqn:line_search}
\text{Inputs}:&&
  w&& = 
  \begin{pmatrix}
    -0.17 & 2.58 & -8.75 \\
    -3.56 & 1.56 & -0.15 \\
     2.15 & -0.66 & 0.49
  \end{pmatrix}, 
 \;\Delta_{init} = 1.0,  N_{iters} = 2\nonumber
\end{aligned}
\end{equation}
\begin{equation}
\begin{aligned} 
\text{$1_{st}$ iteration}: &&
  q = \frac{Q(w,\Delta_{init})}{\Delta_{init}} & =
  \begin{pmatrix}
    0.0 & 3.0 & -7.0 \\
    -4.0 & 2.0 & 0.0 \\
    2.0 & -1.0 & 0.0
  \end{pmatrix} \Rightarrow
 \Delta_{1_{st}} = 1.0 = PO2(\cfrac{91.31}{83.0})\\
\text{$2_{nd}$ iteration}:&&
  q = \frac{Q(w,\Delta_{1_{st}})}{\Delta_{1_{st}}} & = 
  \begin{pmatrix}
    0.0 & 3.0 & -7.0 \\
    -4.0 & 2.0 & 0.0 \\
    2.0 & -1.0 & 0.0
  \end{pmatrix}\Rightarrow
 \Delta_{2_{nd}} = 1.0 = PO2(\cfrac{91.31}{83.0}) \\
\end{aligned}
\label{eqn:MSQE_example}
\end{equation}

\begin{algorithm}
\caption{Line Search}\label{alg:line-search}
\begin{algorithmic}
\Require $w, {\Delta}_{PO2_{init}}, {N}_{range}$
\Ensure ${\Delta_{PO2}^*}$
\Procedure{:}{}
\State $w_q \gets Q(w, {\Delta}_{PO2_{init}})$
\State $msqe^* \gets || w_q - w  ||^2$
\State $\Delta_{PO2}^* \gets {\Delta}_{PO2_{init}}$

\For{\texttt{$offset$ = $-N_{range}$ to ${N}_{range}$}}
\State $\Delta_{PO2} \gets {\Delta}_{PO2_{init}} \cdot 2^{offset}$
\State $w_q \gets Q(w, \Delta_{PO2})$
\State $msqe \gets || w_q - w  ||^2$

\If {\textsc{$msqe < msqe^*$}}:
    \State $msqe^* \gets msqe$
    \State $\Delta_{PO2}^* \gets \Delta$\EndIf
\EndFor
\EndProcedure
\end{algorithmic}
\end{algorithm}

\section{Details on a Problem with Outliers in $\sf{MSQE}$} \label{app:outlier}
\begin{figure}
\centering
  \includegraphics[width=.5\linewidth]{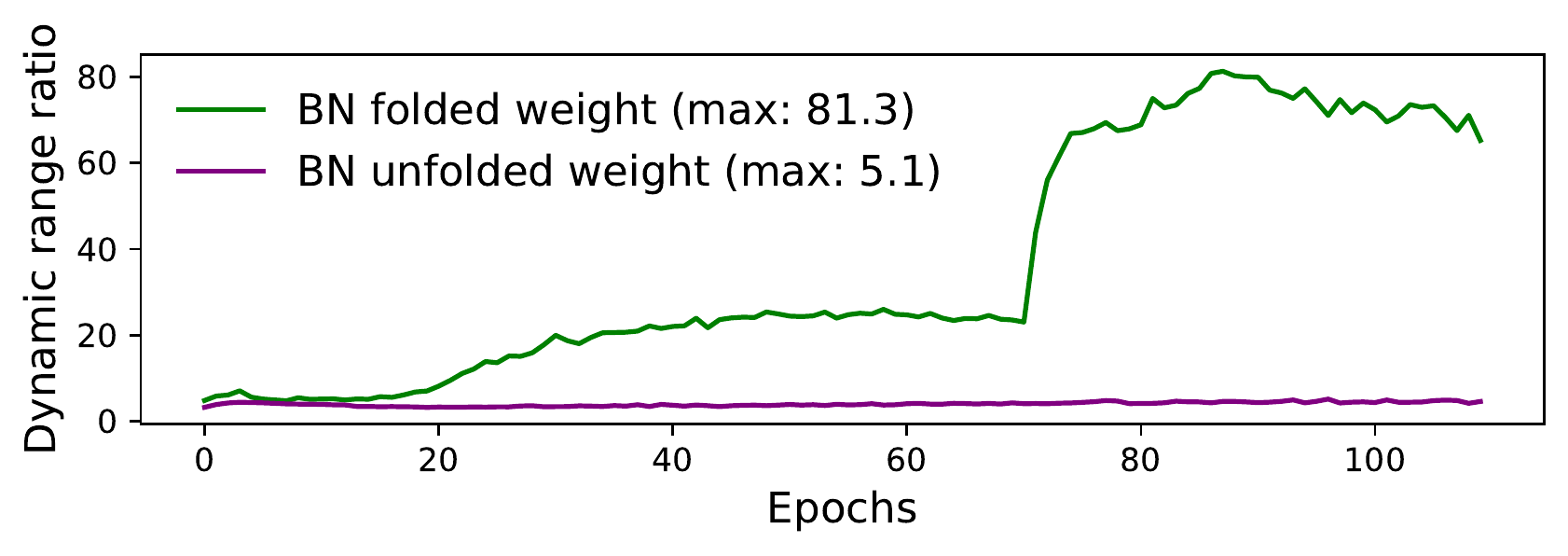}
\caption{It shows that the ratios between the maximum and the minimum dynamic range (within the channels of a layer) for the BN (Batch Normalization) folded weight and the BN unfolded weight, which were averaged across all the layers, from training MobileNetV1 with $\it{MSQE} (finetune)$ quantizers. We can see that the BN folded weight generally had higher ratios compared to the BN unfolded one during training (about 15.9 times higher maximum dynamic range ratio).}
\label{fig:app-max_min_dynamic_range_ratio}
\end{figure}
The equation~\eqref{eqn:app-outlier_mask} (copied here for convenience) shows that a percentile based threshold ($\sigma_{outlier}$) is used to generate an outlier mask for the MSQE optimization to filter out outliers. Each element in the outlier mask gets either 1 or 0 depending on whether its corresponding input value is less than the threshold. Algorithm~\ref{alg:app-weighted-msqe-opt} and~\ref{alg:app-weighted-line-search} show how the outlier mask is applied during the MSQE optimizations. They get the outlier mask as a weight factor ($f_{msqe}$), which is possibly combined (multiplied) with other weight factors, and perform a weighed least squares fit and a weighted line search to ignore the outliers as shown in Algorithm~\ref{alg:app-weighted-msqe-opt} and~\ref{alg:app-weighted-line-search}, respectively.
\begin{equation}    
\begin{aligned}
\label{eqn:app-outlier_mask} 
% \State $M_{outlier} = \mathbbm{1}_{\sigma_{outlier}(w)}(|w|)$ \Comment{\textit{Element-wise 1 or 0, 1 if $w_{i} < T_{outlier}$, and 0 otherwise.}}
M_{outlier} = \begin{cases}  0& \text{if } |w| \geq \sigma_{outlier}\cdot \sigma(w) \\ 
                             1& \text{if } |w| <\sigma_{outlier}\cdot \sigma(w) \end{cases}
\end{aligned}
\end{equation}
\begin{algorithm}
\caption{Weighted MSQE-based Quantizer Optimization}\label{alg:app-weighted-msqe-opt}
\begin{algorithmic}
\Require $w, {\Delta}_{init}, {N}_{iters}, f_{msqe}$
\Ensure ${\Delta_{PO2}}$
\Procedure{:}{}
\State $q \gets \frac{Q(w, \Delta_{init})}{\Delta_{init}}$
\State $q \gets q\odot\sqrt{f_{msqe}}$ \Comment{\textit{Apply the element-wise MSQE weight factor.}}
\State $w \gets w\odot\sqrt{f_{msqe}}$ \Comment{\textit{Apply the element-wise MSQE weight factor.}}
\State $N \gets {N}_{iters}$
\While{$N \neq 0$}
\State $\Delta \gets \cfrac{q^{T}w}{q^{T}q}$ \Comment{\textit{Find an unconstrained optimal $\Delta$ minimizing the MSQE.}}
\State $\Delta_{PO2} \gets PO2(\Delta)$ \Comment{\textit{Constrain $\Delta$ to be PO2.}}
\State $q \gets \frac{Q(w, \Delta_{PO2})}{\Delta_{PO2}}$ \Comment{\textit{Quantize w with the newly found  $\Delta_{PO2}$.}}
\State $q \gets q\odot\sqrt{f_{msqe}}$ \Comment{\textit{Apply the element-wise MSQE weight factor.}}
\State $N \gets N - 1$
\EndWhile
\EndProcedure
\end{algorithmic}
\end{algorithm}
\begin{algorithm}
\caption{Weighted Line Search}\label{alg:app-weighted-line-search}
\begin{algorithmic}
\Require $w, {\Delta}_{PO2_{init}}, {N}_{range}, f_{msqe}$
\Ensure ${\Delta_{PO2}^*}$
\Procedure{:}{}
\State $w_q \gets Q(w, {\Delta}_{PO2_{init}})$
\State $msqe^* \gets || f_{msqe}\odot(w_q - w)  ||^2$ \Comment{\textit{Apply the element-wise MSQE weight factor.}}
\State $\Delta_{PO2}^* \gets {\Delta}_{PO2_{init}}$

\For{\texttt{$offset$ = $-N_{range}$ to ${N}_{range}$}}
\State $\Delta_{PO2} \gets {\Delta}_{PO2_{init}} \cdot 2^{offset}$
\State $w_q \gets Q(w, \Delta_{PO2})$
\State $msqe \gets || f_{msqe}\odot(w_q - w) ||^2$ \Comment{\textit{Apply the element-wise MSQE weight factor.}}

\If {\textsc{$msqe < msqe^*$}}:
    \State $msqe^* \gets msqe$
    \State $\Delta_{PO2}^* \gets \Delta$\EndIf
\EndFor
\EndProcedure
\end{algorithmic}
\end{algorithm}
\section{Details on Gradient Variance Awareness (GVA) Heuristic for Loss-Aware Optimization of $\sf{MSQE}$}\label{sec:app-GVA_opt}
\begin{figure}[h]
    \centering
    \includegraphics[width=.8\linewidth]{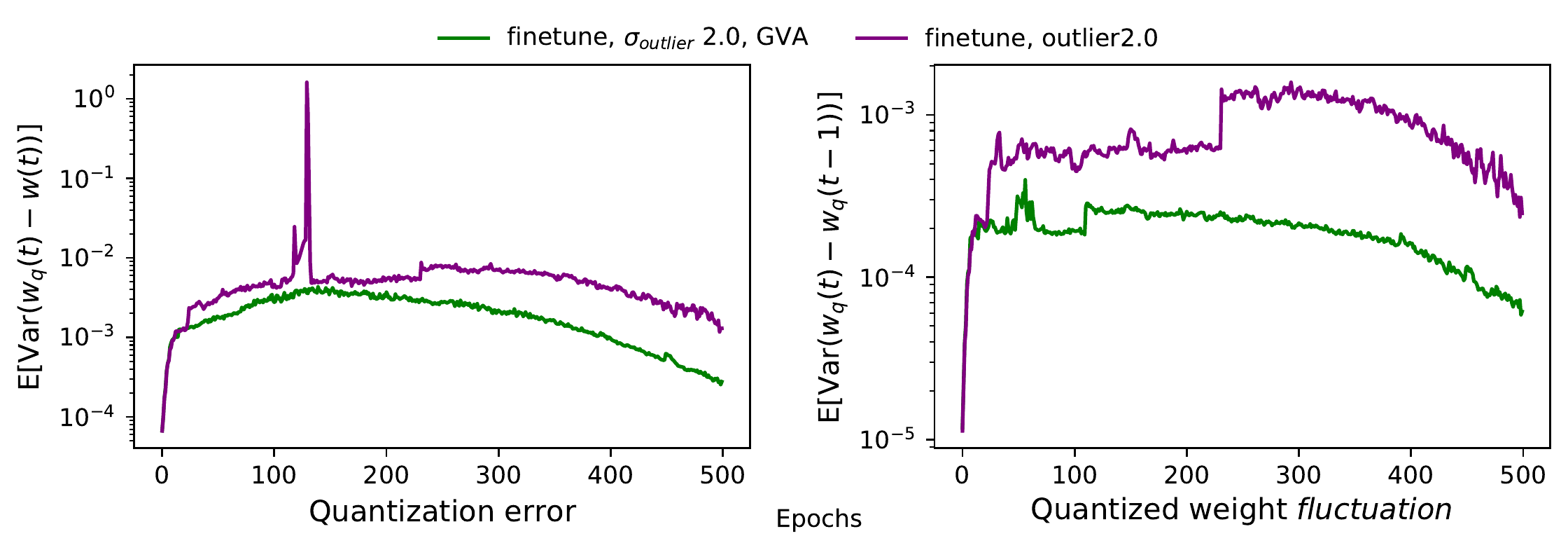}
    \caption[$line_search$]{It shows reduction in the average quantization error variance (left figure) and the average quantized weight $\it{fluctuation}$ variance (right figure) by using the gradient variance aware (GVA) MSQE optimization for MobileNetV2.}
    \label{fig:qnoise_and_fluctuation_mv2}
\end{figure}

The figure~\ref{fig:qnoise_and_fluctuation_mv2} shows reduction in the average quantization error variance (left figure) and the average quantized weight $\it{fluctuation}$ variance (right figure) by using the gradient variance aware (GVA) MSQE optimization for MobileNetV2 (the metrics were measured from the weights and averaged across all the layers).
\section{Details on Convergence Instability of $\sf{GRAD}$}
\subsection{Scaling factor oscillation around the ceil (round) transition boundary}\label{app:conv_instability}
% The figure~\ref{fig:round_to_lower_msqe} shows the effectiveness of the method for a simple example with a single quantizer configured to minimize the L2 error loss of $(w - Q(w))^2$, where the step-size is initialized to near the transition boundary of the ceil function which is also locally MSQE-wise optimum. We can see that the method improves the stability for various random Gaussian noise perturbations to $w$~\footnote{The input $w$ was perturbed by weight element-wise Gaussian noises as $w_{noise} = (1.0  + random\_noise)*w$.}, where the step-size fluctuates when the lower MSQE step-size changes due to the perturbations~\footnote{The average weight perturbations were about 8\% and 16\% of the weight for $\sigma =0.1$ and $\sigma =0.2$ respectively.}.
%

Algorithm~\ref{alg:round_to_lower_msqe} describes the method in detail. It gets the weight ($w$), the unconstrained step-size ($\Delta_{\log_2}$), the running averaged second order moments of the weight ($v_w$), and an outlier mask ($M_{{outlier}^{\Delta_{\log_2}}}$) shown in~\eqref{eqn:app-outlier_mask_grad}. Then it calculates weighted MSQEs of ${\Delta}_{PO2_{\ceil}}$ and ${\Delta}_{PO2_{\floor}}$ to find a PO2 step-size with lower MSQE.
\begin{equation}    
\begin{aligned}
\label{eqn:app-outlier_mask_grad} 
M_{{outlier}^{\Delta_{\log_2}}} = \begin{cases}  0& \text{if } |w| \geq q_{max} \cdot 2^{\Delta_{\log_2}} \\ 
                             1& \text{if } |w| < q_{max} \cdot 2^{\Delta_{\log_2}} \end{cases}
\end{aligned}
\end{equation}

\begin{algorithm}
\caption{RTLM (Round-to-Lower-MSQE)}\label{alg:round_to_lower_msqe}
\begin{algorithmic}
\Require $w, {\Delta_{\log_2}}, v_{w}, M_{{outlier}^{\Delta_{\log_2}}}$
\Ensure ${\Delta_{PO2}^*}$
\Procedure{:}{}
\State ${\Delta}_{PO2_{\ceil}} \gets 2^{\ceil(\Delta_{\log_2})}$
\State ${\Delta}_{PO2_{\floor}} \gets 2^{\floor(\Delta_{\log_2})}$
\State $msqe_{\ceil} \gets || M_{{outlier}^{\Delta_{\log_2}}}\odot v_{w}\odot(Q(w, {\Delta}_{PO2_{\ceil}}) - w)||^2$
\State $msqe_{\floor} \gets || M_{{outlier}^{\Delta_{\log_2}}}\odot v_{w}\odot(Q(w, {\Delta}_{PO2_{\floor}}) - w)||^2$

\If {\textsc{$msqe_{\ceil} < msqe_{\floor}$}}:
    \State $\Delta_{PO2}^* \gets {\Delta}_{PO2_{\ceil}}$
\Else:
    \State $\Delta_{PO2}^* \gets {\Delta}_{PO2_{\floor}}$
\EndIf
\EndProcedure
\end{algorithmic}
\end{algorithm}
\subsection{Scaling factor oscillation at convergence} \label{app:oscillation_at_convergence}
Algorithm~\ref{alg:app-running_ave_po2_step_size} describes the freezing method to improve the scaling factor instability at convergence. It estimates a running average of the PO2 scaling factor exponent ($EMA_{\log_2\Delta_{\potwo}}$) and returns the PO2 scaling factor based on $EMA_{\log_2\Delta_{\potwo}}$ only when the indicator ($\mathbbm{1}_{freeze}$) value is set true.
\begin{algorithm}
\caption{Running-Averaged PO2 Step-Size}\label{alg:app-running_ave_po2_step_size}
\begin{algorithmic}
\Require  $\Delta_{\potwo}, \mathbbm{1}_{freeze}$
\Ensure $\Delta_{\potwo}$
\Procedure{:}{}
\If {\textsc{$\mathbbm{1}_{freeze}$}}:
    \State $\beta \gets 1.0$
\EndIf

\State $EMA_{\log_2\Delta_{\potwo}}\gets \beta\cdot EMA_{\log_2\Delta_{\potwo}} + (1.0 - \beta)\cdot  log_2(\Delta_{\potwo})$

\If {\textsc{$\mathbbm{1}_{freeze}$}}:
    \State $\Delta_{\potwo} \gets 2^{\round(EMA_{\log_2\Delta_{\potwo}})}$
\Else:
    \State $\Delta_{\potwo} \gets \Delta_{\potwo}$
\EndIf
\EndProcedure
\end{algorithmic}
\end{algorithm}
%
% \section{Additional figures}

% As shown in the figure~\ref{fig:scale_fluctuation_5layers}, the step-size fluctuated at near convergence for $\sf{GRAD}$~\footnote{The learning rate was about 2.45e-4 at the 450th epoch.}. 
% \begin{figure}[h]
%     \centering
%     \includegraphics[scale=0.23]{figs/experiments/mv1_scale_fluctuation_5_layer_450.pdf}
%     \caption[$round-to-lower-msqe$]{It shows step-size exponent fluctuations (per training step) of Round-to-Lower-MSQE and Ceil functions for the first 5 layers of MobileNetV1 for 3 epochs from the 450th epoch.}
%     \label{fig:scale_fluctuation_5layers}
% \end{figure}

\section{Extended Results}
\subsection{Evaluation of the Gradient Variance Awareness (GVA) Heuristic for $\sf{MSQE}$} \label{sec:app-results-GVA}
\begin{figure}[ht]
    \centering
    \includegraphics[scale=0.43]{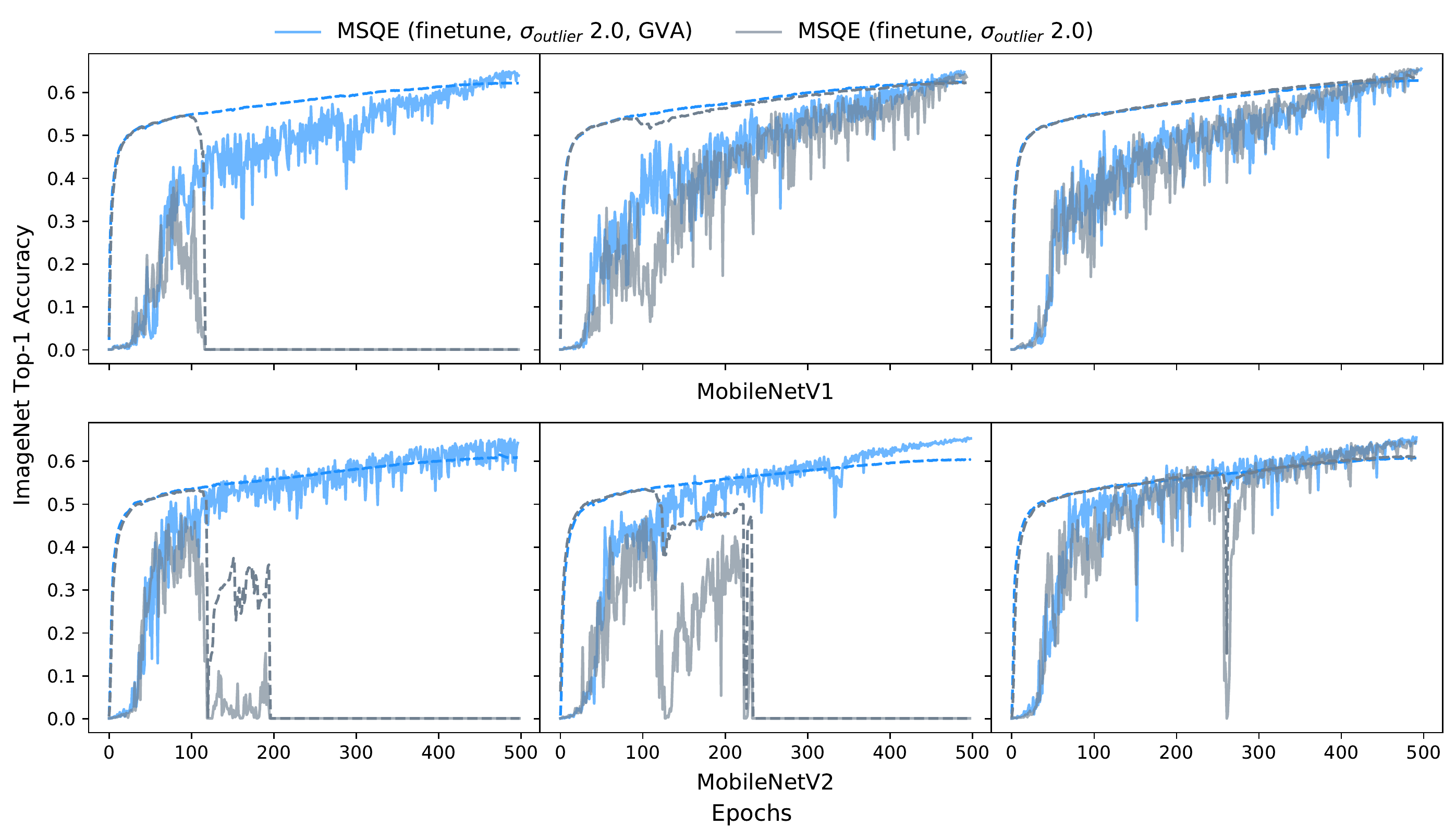}
    \caption[$scaling factor-fluctuation$]{It shows all the three training curves for each of MobileNetV1 (top three figures) and MobileNetV2 (bottom three figures) with and without the GVA method. We can see that the outlier mask method alone ($\sf{MSQE} (finetune, \sigma_{outlier} 2.0)$) resulted in training instability for MobileNetV1 and MobileNetV2.}
    \label{fig:app-outlier_gva}
\end{figure}

\begin{figure}[h]
    \centering
    \includegraphics[scale=0.43]{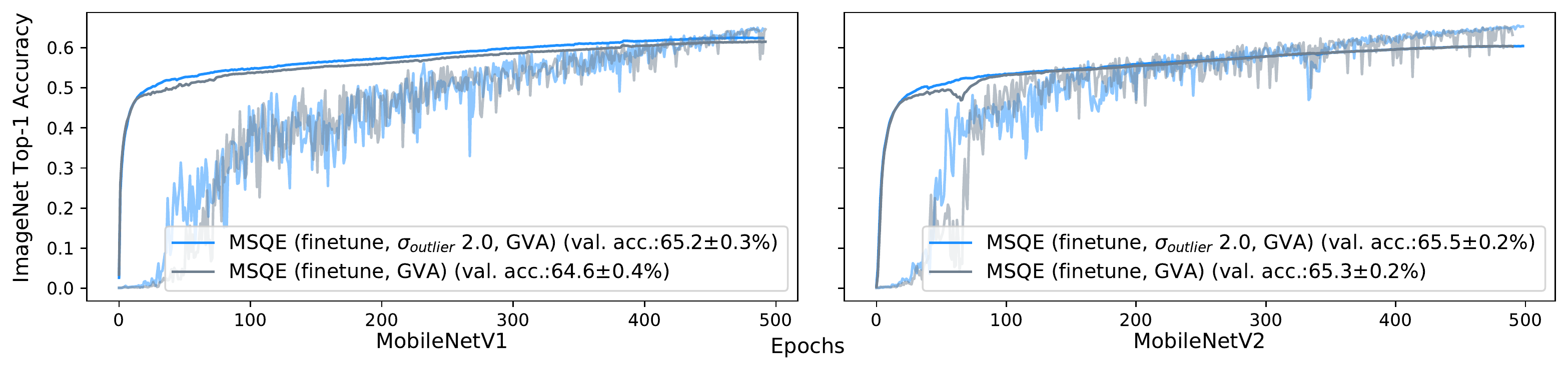}
    \caption[$scaling factor-fluctuation$]{Adding the outlier method to $\sf{MSQE} (finetune, GVA)$ improved training stability in the early epochs and the validation accuracy.}
    \label{fig:ablation_outlier_gva}
\end{figure}

We performed an ablation study to see how much the outlier method has an impact on the overall performance in the experiment in Section~\ref{sec:gva_results}. As we can see from the figure~\ref{fig:ablation_outlier_gva}, adding the outlier method to $\sf{MSQE} (finetune, GVA)$ (the GVA only method) improved training stability in the early epochs and the validation accuracy.

\subsection{Evaluation of RTLM of $\sf{GRAD}$} \label{sec:app-results-RTLM}

\begin{figure}[ht]
    \centering
    \includegraphics[scale=0.37]{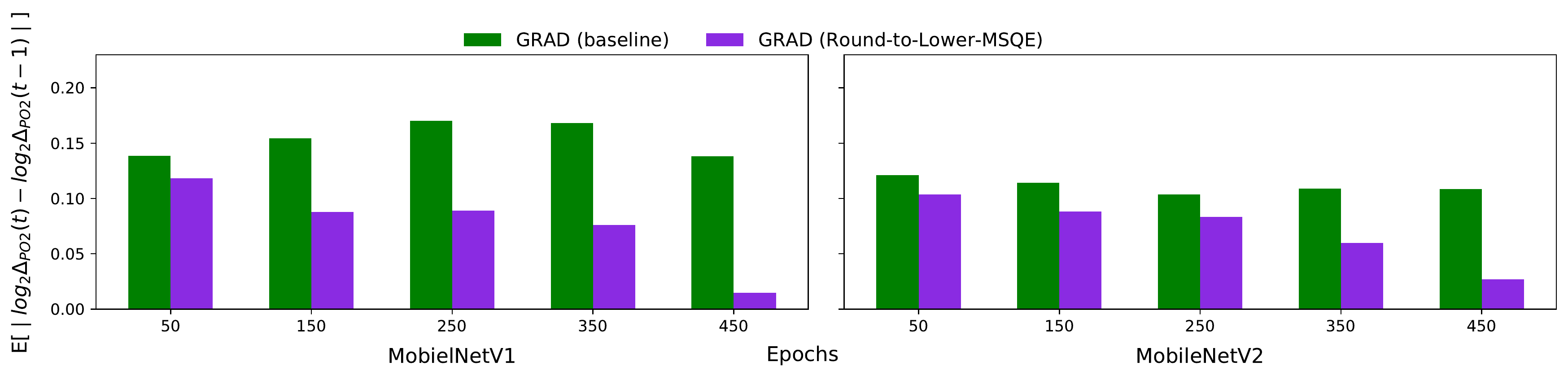}
    \caption[$round-to-lower-msqe$]{It shows average scaling factor exponent fluctuations (per training step) for training MobileNetV1 and MobileNetV2 for 3 epochs from the corresponding epochs on the x-axis. We can see that RTLM had less average scaling factor exponent fluctuations compared to the ceil function.}
    \label{fig:average_scale_fluctuation}
\end{figure}